\documentclass[letterpaper, 10 pt, conference]{ieeeconf}  

\IEEEoverridecommandlockouts                              

\usepackage{amsmath}
\usepackage{amssymb}
\usepackage[
    colorlinks=true,
    linkcolor=black,
    citecolor=black,
    urlcolor=blue
]{hyperref}
\usepackage{cleveref}
\usepackage{xcolor}
\usepackage{amsmath,amsfonts,bm,bbm,mathtools,amssymb}
\usepackage{physics}
\crefname{algocf}{algorithm}{algorithms}
\Crefname{algocf}{Algorithm}{Algorithms}
\usepackage{csquotes}
\usepackage{tikz}

\usepackage{booktabs}

\usepackage{wrapfig}

\usepackage{etoolbox}
\AtBeginEnvironment{appendices}{\crefalias{section}{appendix}}

\def\eqref#1{equation~\ref{#1}}

\def\1{\bm{1}}

\def\vxi{{\bm{\xi}}}
\def\vtau{{\bm{\tau}}}
\def\vxi{{\bm{\xi}}}

\def\vc{{\bm{c}}}

\DeclareMathAlphabet{\mathsfit}{\encodingdefault}{\sfdefault}{m}{sl}
\SetMathAlphabet{\mathsfit}{bold}{\encodingdefault}{\sfdefault}{bx}{n}

\usepackage{bbding}

\usepackage{graphicx}      
\usepackage{tikz}
\usepackage{pgfplots}
\pgfplotsset{compat=newest}
\usetikzlibrary{plotmarks}
\usetikzlibrary{arrows.meta}
\usepgfplotslibrary{patchplots}
\usepackage{grffile}

\usepackage{xcolor}
\usepackage{xfrac}
\usepackage{bm}
\usepackage{mathtools}
\usepackage{multicol}
\usepackage{siunitx}
\usepackage{multirow}
\usepackage{tabularx}
\usepackage{stackengine}
\usepackage{comment}
\usepackage[short, nocomma]{optidef}
\usepackage{algorithm}
\usepackage{algpseudocode}
\usepackage{multicol}
\usepackage{microtype}
\DeclareSIUnit{\rad}{rad}

\usepackage[
    style=ieee,
    backend=biber,
    doi=false,
    url=false,
    isbn=false,
    eprint=false,
    maxnames=20,
    minnames=2,
    sorting=none,
    mcite=true,
    sortcites=true,  
    citestyle=numeric-comp,  
]{biblatex}

\AtBeginBibliography{\footnotesize} 
\newcommand{\citet}{\cite}

\newcommand{\R}[1]{\ensuremath{\mathbb{R}^{#1}}}

\newcommand{\N}{\ensuremath{\mathcal{N}}}

\DeclareMathAlphabet{\mymathbb}{U}{BOONDOX-ds}{m}{n}

\newcommand{\upright}[2]{\ensuremath{#1_{\mathrm{#2}}}}

\newcommand{\dimy}{\ensuremath{\upright{n}{y}}}
\newcommand{\dimxi}{\ensuremath{\upright{n}{\xi}}}

\colorlet{lightgreen}{green!80!black}
\colorlet{darkgreen}{green!50!black}
\colorlet{lightblue}{cyan!60!white}
\definecolor{orange_1}{RGB}{254, 163, 68}   
\definecolor{blue_1}{RGB}{199, 217, 231}    
\definecolor{red_1}{RGB}{242, 154, 154}    

\DeclareRobustCommand{\graycircle}{\tikz[baseline=-0.75ex]\draw[gray,fill=gray] (0,0) circle (.5ex);}
\DeclareRobustCommand{\greencircle}{\tikz[baseline=-0.75ex]\draw[darkgreen,fill=darkgreen] (0,0) circle (.5ex);}
\DeclareRobustCommand{\redcircle}{\tikz[baseline=-0.75ex]\draw[red,fill=red] (0,0) circle (.5ex);}
\DeclareRobustCommand{\lightredcircle}{\tikz[baseline=-0.75ex]\draw[red_1,fill=red_1] (0,0) circle (.5ex);}
\DeclareRobustCommand{\bluecircle}{\tikz[baseline=-0.75ex]\draw[blue_1,fill=blue_1] (0,0) circle (.5ex);}
\DeclareRobustCommand{\redboundcircle}{\tikz[baseline=-0.75ex]\draw[red, thick] (0,0) -- (1.3em,0);}

\DeclareRobustCommand{\solidlightgreen}{\tikz[baseline=-0.4ex]\draw[lightgreen, thick] (0,0) -- (1.3em,0);}
\DeclareRobustCommand{\dashedblack}{\tikz[baseline=-0.5ex]\draw[black, dashed, thick] (0,0) -- (1.3em,0);}
\DeclareRobustCommand{\soliddarkgreen}{\tikz[baseline=-0.5ex]\draw[darkgreen, thick] (0,0) -- (1.3em,0);}
\DeclareRobustCommand{\dashedlightblue}{\tikz[baseline=-0.5ex]\draw[lightblue, dashed, thick] (0,0) -- (1.3em,0);}
\DeclareRobustCommand{\solidyellow}{\tikz[baseline=-0.5ex]\draw[orange_1, thick] (0,0) -- (1.3em,0);} 
\DeclareRobustCommand{\dashedmagenta}{\tikz[baseline=-0.5ex]\draw[magenta, dashed, thick] (0,0) -- (1.3em,0);}
\DeclareRobustCommand{\dashedgray}{\tikz[baseline=-0.5ex]\draw[gray, dashed, thick] (0,0) -- (1.3em,0);}

\title{\LARGE \bf
Integrating Flow Matching and Model Predictive Control\\ for Safe Robot Navigation
}

 \author{ Philipp Holzmann$^{1}$, Maik Pfefferkorn$^{1}$, Jo\~{a}o Carvalho$^{2}$, Ali Younes$^{3}$, An T. Le$^{2}$,\\ Georgia Chalvatzaki$^{3,5,6}$, Jan Peters$^{2,4,5,6}$, Rolf Findeisen$^{1}$
 \thanks{$^{1}$P. Holzmann, M. Pfefferkorn and R. Findeisen are with the Control and Cyber-Physical Systems Laboratory,
         Technical University of Darmstadt, Germany
         {\tt\small \{givenname.surname\}@iat.tu-darmstadt.de}}%
 \thanks{$^{2}$J. Carvalho, A. T. Le and J. Peters are with the Intelligent Autonomous Systems Laboratory, Technical University of Darmstadt, Germany
          {\tt\small \{givenname.surname\}@tu-darmstadt.de}}%
 \thanks{$^{3}$A. Younes and G. Chalvatzaki are with the Interactive Robot Perception and Learning Laboratory, Technical University of Darmstadt, Germany
         {\tt\small \{givenname.surname\}@tu-darmstadt.de}}%
 \thanks{$^4$German Research Center for AI (DFKI), Research Department SAIROL}%
 \thanks{$^5$Hessian Centre for Artificial Intelligence}%
 \thanks{$^6$Robotics Institute Germany (RIG)}%
 }%

\allowdisplaybreaks

\begin{document}

\maketitle
\thispagestyle{empty}
\pagestyle{empty}

\begin{abstract}
Learning-based path planners based on diffusion and flow matching can generate diverse trajectories from demonstrations but typically lack mechanisms for constraint satisfaction and collision avoidance during deployment. 
This work proposes a framework that combines flow-matching-based path planning with model predictive path-following control to integrate data-driven path generation with real-time, constraint-aware execution. 
The flow-matching model captures multimodal path distributions, while the predictive controller adapts motion online to satisfy state and input constraints and avoid obstacles using updated local sensor information. 
An event-triggered re-planning scheme incorporates predictive control solutions into subsequent path generation, promoting collision-free navigation in environments with previously unseen obstacles. 
Simulation results for mobile robot navigation demonstrate robust performance across cluttered environments.
The framework provides a systematic approach to combining flow matching with predictive control for adaptive, data-driven navigation.
\end{abstract}

\section{Introduction}
Recent advances in robot control for complex tasks have been driven by increasingly capable machine learning algorithms and the availability of high-performance computing resources~\cite{DBLP:journals/jmlr/LevineFDA16,DBLP:conf/rss/HaarnojaHZTTL19,chi2024}.
Machine learning enables expressive control policies that often outperform classical methods in high-dimensional and multimodal settings, particularly when trained via imitation or reinforcement learning.

To handle multimodal data distributions, recent work has applied deep generative models based on diffusion~\cite{ho2020denoisingdiffusion} and flow matching~\cite{Lipman2023,Lipman2024}, originally developed for image synthesis, to robotics domains including imitation learning~\cite{DBLP:conf/rss/ChiFDXCBS23,DBLP:conf/rss/ReussLJL23}, offline RL~\cite{janner2022diffuser}, grasping~\cite{urain2022se3diffusion,DBLP:journals/corr/abs-2412-08398}, and motion planning~\cite{carvalho2022conditionedsbm,DBLP:conf/iros/Carvalho0BK023,carvalho2025motion,Liu2024,Nguyen2025,Dai2025,Braun2024}. These models are particularly attractive for their ability to represent complex, multimodal trajectory distributions and to generate diverse candidate motions from demonstration data, including different paths from the same initial point.
In this article, we use a variant of flow matching the shortcut model~\cite{Frans2024}, which offers fast sampling~\cite{Schusterbauer2025}.

One central challenge in learning policies from demonstrations is the lack of guarantees for collision avoidance during deployment. 
For example, a generative model trained on human-demonstrated trajectories may produce paths that become unsafe when new obstacles are introduced or when dynamics or constraints (e.g., velocity limits) change. 
Nevertheless, the data remains valuable as a reference for a lower-level controller that can safely adapt to it.

\begin{figure*}[thb]
\centering
\includegraphics[width=\textwidth, trim={0.0cm 0cm 0.0cm 0.3cm},clip]{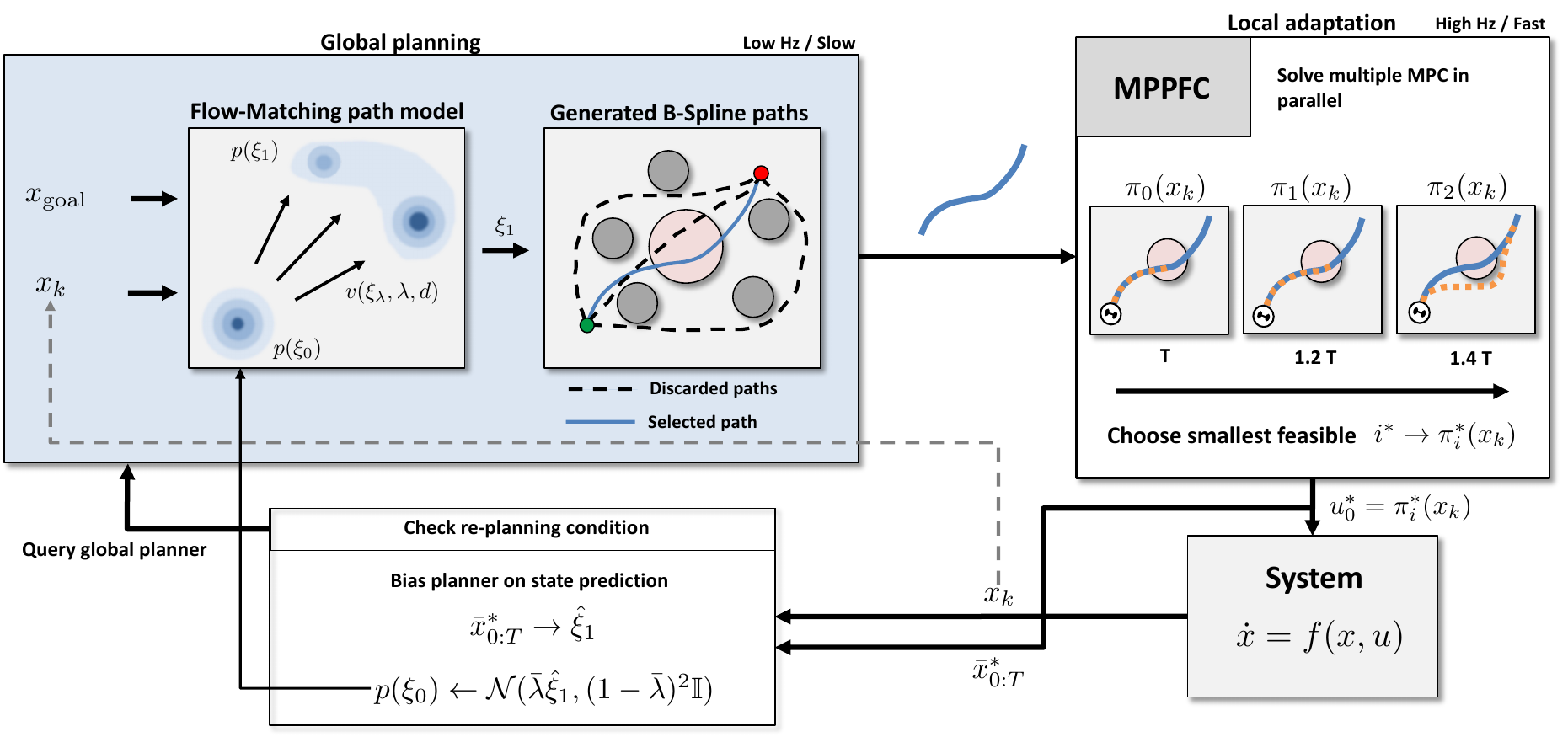}\vspace{-5mm}
\caption{Global flow-matching-based path planning with local adaptation via model predictive path-following control (MPPFC). Given the current and goal system states, a batch of path parametrizations is sampled from the generative model. A collision-free path, considering known obstacles (gray circles), is selected (in blue), and multiple MPPFC instances with varying prediction horizons optimize the system inputs in parallel to follow it. If unforeseen obstacles (light red circle) make the current path or optimal control problem infeasible, the MPC with the shortest feasible prediction horizon is chosen and applied to the system. Once a specific condition is met, replanning is triggered to determine a new global plan biased by the latest state prediction.}\vspace{-3mm}\label{fig:scheme}
\end{figure*}

To this end, we combine the deep generative planner with a model predictive path-following controller (MPPFC)~\cite{matschek2019nonlinear,faulwasser2015nonlinear,Faulwasser2009}, which tracks and locally reshapes the generated path to ensure safety, cf. Fig.~\ref{fig:scheme}.
Unlike tracking model predictive control (MPC)~\cite{Rawlings2019,Findeisen2007}, MPPFC treats path progression and velocity as optimization variables, enabling adaptive motion profiles and reactive behavior in dynamic environments. 
By enforcing input and state constraints, including collision avoidance and actuator limits, MPPFC ensures constraint satisfaction even if the reference path must be temporarily deviated from.

Previous works have used simpler generative models, focused on unimodal settings and control-theoretic guarantees~\cite{Matschek2020a,Matschek2020b,Matschek2021,Holzmann2022,Pfefferkorn2022}.
Since combining diffusion and flow matching generative models with predictive control offers a large potential~\cite{Huang2024}, this has been explored by a series of works: \citet{Zhou2025} uses a diffusion model to learn multi-step predictions and warm-start MPC optimizations;~\citet{Julbe2025} parameterizes MPC feedback using diffusion;~\citet{Huang2025} applies diffusion models to solve nonlinear MPC problems; and~\citet{Xue2024,kurtz2025generative} explore diffusion rollouts for model predictive path integral control (MPPIC). 
Relatedly,~\citet{Huang2024} uses diffusion to generate subgoals for long-horizon control with MPPIC.
However, because of the underlying MPC methods used, these works do not guarantee safety or path-following.
In contrast, our approach combines flow-matching for reference generation with MPPFC for local safety and reference tracking, overcoming the limitations of the current methods.
By using MPPFC to track and refine paths generated via flow matching, we ensure constraint satisfaction --- particularly collision avoidance --- and enable real-time adaptation. 
In this sense, the MPPFC shows similarities to predictive safety filters~\cite{Milios2023,Wabersich2021,Wabersich2023}.  
However, in contrast to standard safety filters that modify unsafe inputs to ensure safety, we directly apply MPPFC as the main controller, using its structure to dynamically shape both path execution and progression.

The main contributions of this work are: 
(1) a framework that combines a flow-matching-based path generator, learned from human demonstrations, with MPPFC to enable constraint satisfaction and collision avoidance during execution;
(2) a bidirectional coupling in which flow-generated paths guide MPPFC and its predictions bias the flow planner during event-triggered re-planning;
(3) a systematic evaluation demonstrating collision-free path-following in environments with multiple obstacle configurations not observed during training.

The remainder of the paper is structured as follows: Section~\ref{sec:ProblemFormulation} outlines the problem setting. Section~\ref{sec:GenerativeModeling} introduces the generative model used to learn path distributions from demonstrations. Section~\ref{sec:MPC} details the MPPFC formulation and its integration with the planner. Section~\ref{sec:Simulation} presents simulation results, and Section~\ref{sec:Conclusions} concludes the paper.

\section{Problem Formulation}\label{sec:ProblemFormulation}
Consider a dynamical system governed by known continuous-time dynamics:
\begin{subequations}\label{eq:general_dyn}
\begin{align}
\Dot{x}(t) &= f(x(t), u(t)), \quad x(t_0) = x_0, \\
y(t) &= h(x(t)).
\end{align}
\end{subequations}
Here, $x \in \mathcal{X} \subseteq \mathbb{R}^{n_x}$ represents the system state, $u \in \mathcal{U} \subseteq \mathbb{R}^{n_u}$ is the control input, and $y \in \mathcal{Y} \subseteq \mathbb{R}^{n_y}$ denotes the system output. 
The function $f: \mathbb{R}^{n_x} \times \mathbb{R}^{n_u} \to \mathbb{R}^{n_x}$ defines the system dynamics, while $h: \mathbb{R}^{n_x} \to \mathbb{R}^{n_y}$ defines the output map. 
The initial state at time $t_0$ is given by $x_0 \in \mathbb{R}^{n_x}$.

Given a global reference path, the control objective is to navigate the system through a cluttered environment while strictly satisfying system constraints and avoiding collisions.
Rather than relying on classical sampling-based planners (e.g., RRT~\cite{kuffner2000rrtconnect}), we assume access to expert task demonstrations that encode nuanced human preferences for spatial navigation. To model and reproduce these demonstrations, we learn a flow-matching-based generative planner.

Given start and goal configurations, candidate paths are generated by sampling from this learned flow-matching model. 
Formally, a path is defined as a geometric curve $\rho: \mathcal{C}_\theta \to \mathcal{Y}, \theta \mapsto y=\rho(\theta)$ in the system's output space, parametrized by a scalar path parameter $\theta$. 
We represent each path as a linear combination of a finite set of fixed-order B-splines, training the flow-matching model directly on the governing coefficients.

While the generative planner provides a strong, obstacle-aware prior based on demonstrations, it is not strictly tasked with enforcing real-time constraints or active collision avoidance using all current local information. 
Instead, this responsibility is systematically delegated to a lower-level controller that incorporates this information.

Specifically, the generated paths serve as reference priors for parallel instances of an MPPFC, each configured with a different prediction horizon. 
The predictive controller seamlessly enforces state and input constraints, adapting the trajectory online to navigate environmental changes and previously unseen obstacles. During closed-loop operation, the generative model is selectively resampled whenever predefined criteria are met (e.g., a significant spatial deviation from the reference path). 
Crucially, these newly generated paths are biased by the predictive controller's solution trajectory. 
This bidirectional coupling steers the generative model toward inherently collision-free regions of the task space, ensuring tight consistency between global planning and local execution while minimizing the corrective effort demanded of the controller.

\section{Generative Modeling for Path Planning}
\label{sec:GenerativeModeling}

We begin by introducing the path parametrization used to represent output paths.
We then describe how demonstration data are converted into this parametrized representation for use in learning.
Finally, we provide an overview of the generative model of expert behaviors, formulated as a probabilistic model over paths inferred from demonstrations.

\subsection{From Trajectory Data to Path Parametrization}

We parameterize the output path $\rho(\theta) \in \mathbb{R}^{\dimy}$ as a linear combination of $M\in \mathbb{N}^*$ clamped B-splines of order $r\in \mathbb{N}$. For each dimension $j$, the path is given by
\begin{equation}\label{eq:path_func}
    \rho_j(\theta) = \sum_{m=1}^M \xi_{m,j} B_{m, r}(\theta),
\end{equation}
where $\xi_{m,j}$ are the governing coefficients and $B_{m,r}(\theta)$ is defined via the Cox-de Boor recursion formula~\cite{DeBoor1976}. 
The full path $\rho(\theta)$ is thus characterized by $M \dimy$ coefficients.

Given $N\in \mathbb{N}^*$ time-indexed expert demonstrations $\vtau^{(i)} = \{ (t_k, y(t_k)) \}_{k=0}^{n_i}$, we convert these into time-independent geometric paths. 
This representation is agnostic to variations in trajectory lengths, sampling rates, and velocity profiles.
We extract the corresponding coefficient dataset $\mathcal{D}_\xi = \{ \vxi^{(i)} \}_{i=1}^N$ from the trajectory data $\mathcal{D}_\tau = \{\vtau^{(i)} \}_{i=1}^N$ by solving a linear least-squares regression for each demonstration as
\begin{align}\label{eq:fit_para}
    \vxi^{(i)} \! = \! \arg\min_{\vxi^{(i)}} \! \Bigg \{ \! \sum_{j=0}^{n_i} \! \| y(t_j) \! - \! \rho(\theta_j) \|^2 \Big \vert (t_j, y(t_j)) \! \in \! \vtau^{(i)} \! \! \Bigg \},
\end{align}
using evaluation points $\theta_j = (t_j - t_0)/(t_{n_i} - t_0)$. 
This minimization has a standard closed-form solution~\cite{bishop2006pattern}.

\subsection{Modeling and Learning Expert Behaviors}

To model expert behavior, we learn a shortcut model~\cite{Frans2024}, a variant of flow matching, on the dataset $\mathcal{D}_\xi$.
This model encodes the distribution of path coefficients $p(\vxi | \vc)$, conditioned on a context $\vc \in \mathbb{R}^{n_\vc}$, representing the start and goal states, and captures the (potentially multi-modal) distribution of expert behaviors in the environment.
This distribution is modeled with flow-matching using a time-varying vector field whose solutions map samples from an initial noise distribution to the data distribution.
The shortcurt model is trained to match the displacement between noise and demonstration samples while enforcing consistency across integration steps, using the loss function
\begin{multline}
    L_\zeta = \mathbb{E}_{\vxi_0, \vxi_1, (\lambda,\delta)} \Big[ \| v_\zeta(\lambda, \vxi_\lambda, 0) - (\vxi_1 - \vxi_0)\|^2 \\
    + \|v_\zeta(\lambda, \vxi_\lambda, \delta) - \bar{v}\|^2 \Big]. 
\end{multline}
Here, $\vxi_0 \sim \N(0, \mathbb{I})$, $\vxi_1 \sim \mathcal{D}_\xi$, and $(\lambda,\delta) \sim \mathcal{U}(0,1)\times\mathcal{U}(0,1)$.
Furthermore, $\vxi_\lambda = (1-\lambda)\vxi_0 + \lambda \vxi_1$, and $\bar{v} = 0.5[v_\zeta(\lambda, \vxi_\lambda, 0.5 \delta) + v_\zeta(\lambda+\delta, \vxi_{\lambda+\delta}, 0.5 \delta)]$. 
The resulting model can efficiently generate expert-like path coefficients, and we refer the reader to \cite{Frans2024} for full details.

\section{Model Predictive Control for Path Following and Safety Filtering}
\label{sec:MPC}

Once a path is generated using the expert model, we use it as a reference in a MPPFC. 
We first review the basics of MPPFC and then discuss its application as an integrated local safety filter and tracking controller, ensuring collision-free closed-loop behavior.
Finally, we detail the proposed integration of the learned path planner with the MPPFC.

\subsection{Model Predictive Path-Following Control}
We use a MPPFC to steer the system (\ref{eq:general_dyn}) along a given path, following established approaches~\cite{matschek2019nonlinear,faulwasser2015nonlinear,Faulwasser2009}. 
To enable parallel evaluation of multiple horizon lengths, we consider $K\in\mathbb{N}$ controllers, indexed by $i \in \{0, 1, \dots, K-1\}$.
Each controller employs an extended prediction horizon of $(T + i\delta_t)$, where  $T \in\mathbb{R}_{>0}$ denotes a base horizon length and $\delta_t  \in\mathbb{R}_{>0}$ is a fixed increment that enables systematical exploration of longer horizons. 
The corresponding optimization problem is formulated as
%
\begin{subequations}\label{eq:OCP}
\begin{alignat}{2}
    \pi_i(x_k) = & \arg & & \min_{\Bar{u},\Bar{w}} J\big(\Bar{e},\Bar{u}, \theta, \Bar{w} \big) \label{eq:OCP_cost} \\
    & ~ \text{s.~t.} & & ~~ \forall s \in [0, T+ i \delta_t]: \notag \\
    &  & & ~~ ~~~~ \Dot{\Bar{x}}(s) = f(\Bar{x}(s),\Bar{u}(s)), ~ \Bar{x}(0) = x_k \label{eq:OCP_sysdyn} \\
    &  & & ~~ ~~~~ \Dot{\theta}(s) = g(\theta(s), \bar{w}(s)), ~ \theta(0)= \theta_k \label{eq:OCP_pathdyn} \\
    &  & & ~~ ~~~~ \Bar{y}(s) = h(\Bar{x}(s)) \label{eq:OCP_sysout} \\
    &  & & ~~ ~~~~ \Bar{e}(s) = \rho(\theta(s)) - \Bar{y}(s) \label{eq:OCP_error} \\
    &  & & ~~ ~~~~ \theta(s) \in \mathcal{C}_{\theta}, ~ \Bar{w}(s) \in \mathcal{W} \label{eq:OCP_pathcons} \\
    &  & & ~~ ~~~~ \Bar{x}(s) \in \mathcal{X}_\mathrm{safe} \subseteq \mathcal{X}, ~ \Bar{u}(s) \in \mathcal{U} \label{eq:OCP_syscons} \\
    &  & & ~~ [ \Bar{x}^\top(T+ i\delta_t) \;\; \theta^\top(T+ i\delta_t)]^\top \in \Omega. \label{eq:OCP_termcon} 
\end{alignat}
\end{subequations}
The controller $\pi_i$ predicts the system's future behavior using (\ref{eq:OCP_sysdyn}) and (\ref{eq:OCP_sysout}) to optimize the future input trajectory $\bar{u}$, typically parameterized as piecewise constant.
Physical limits and collision-avoidance requirements are captured by the system state and input constraint sets $\mathcal{X}_\mathrm{safe}$ and $\mathcal{U}$ in (\ref{eq:OCP_syscons}).
Concurrently, the virtual dynamics (\ref{eq:OCP_pathdyn}) and virtual input $\bar{w}$ control the progress along the reference path, effectively coupling planning and control. 
The corresponding constraints in (\ref{eq:OCP_pathcons}) act as tuning parameters.
The terminal cost (\ref{eq:OCP_termcon}) ensures consistency between the true and the virtual states at the end of the horizon, promoting convergence to the reference path while respecting system and path constraints.

The cost function (\ref{eq:OCP_cost}) is defined as
\begin{equation}
    \begin{split}
        J\big(\Bar{e},\Bar{u},\theta, \Bar{w}\big) = \int_{0}^{T+ i\delta_t} & L(\Bar{e}(s),\Bar{u}(s),\theta(s), \Bar{w}(s)\big) \dd s \\  &+ E(\Bar{x}(T+ i\delta_t),\theta(T+ i\delta_t)\big), 
    \end{split}
\end{equation}
where the stage cost $L$ penalizes path deviations (\ref{eq:OCP_error}), large control inputs, and the distance of $\theta$ to its final value, encouraging forward progress. The terminal cost $E$ and constraint (\ref{eq:OCP_termcon}) jointly ensure convergence to the path when chosen appropriately~\cite{faulwasser2015nonlinear}.
Problem (\ref{eq:OCP}) can be solved with gradient-based (gMPPFC) or sampling-based (sMPPFC)~\cite{Williams2000, Theodorou2010pathintegral} methods, though the latter do not guarantee strict constraint satisfaction.

\subsection{MPPFC for Safety Filtering}
\label{sec:safety_filter}
\begin{figure*}[thb]
    \centering
    \includegraphics[width=\textwidth, trim={0.3cm 0.2cm 0.32cm 0.4cm},clip]{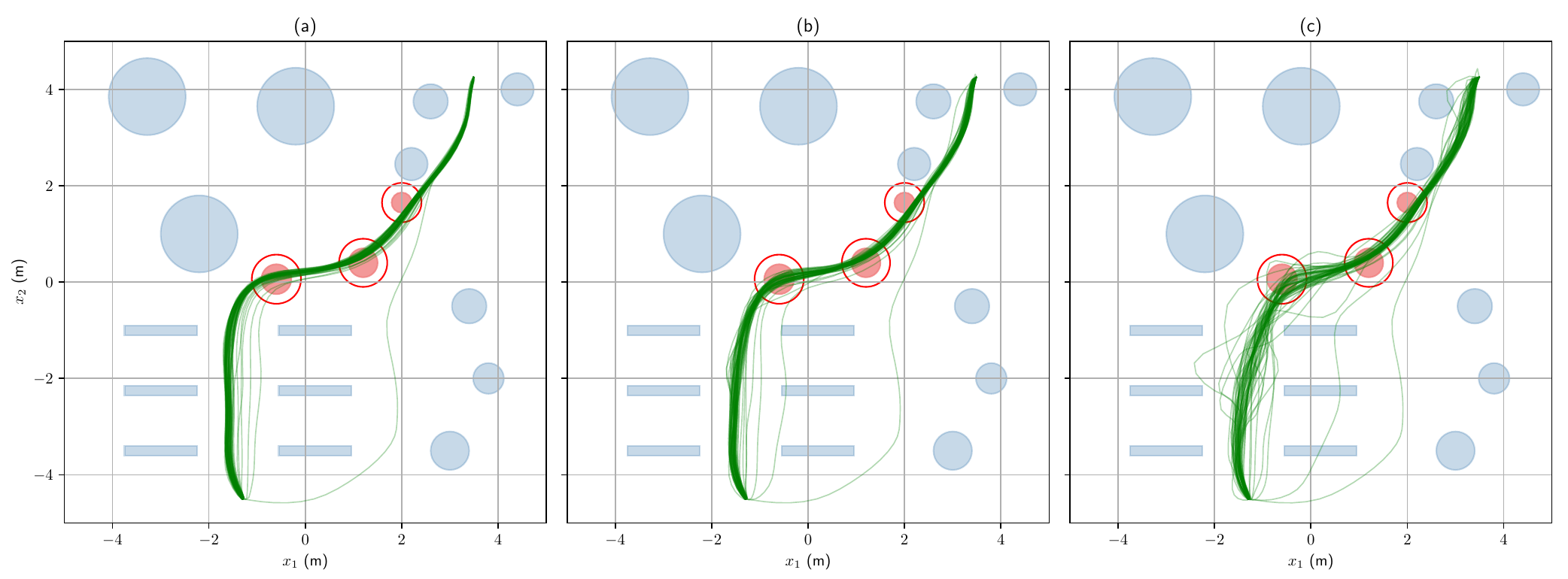}
    \vspace{-5mm}
    \caption{Path planning using a flow matching model guided by an additive gradient. Blue indicates obstacles present during data collection, whereas new obstacles are depicted in light red, with the red circle denoting a clearance margin for the mobile robot's footprint. Generated path samples are plotted as green lines. The subfigures demonstrate: (a) unguided sampling, (b) guidance with a low scale ($\gamma_\mathrm{g} = 0.2$), and (c) guidance with a high scale ($\gamma_\mathrm{g} = 0.6$).}\vspace{-3mm}
    \label{fig:guidance}
\end{figure*}

The flow matching model generates reference paths that provide a strong, obstacle-aware prior based on the demonstrations. To systematically address dynamic environmental changes, such as new or unforeseen obstacles, we integrate this generative planning with MPPFC using up-to-date local sensor information. 

In principle, one could adapt the generative model directly to new obstacles using an energy-based guidance term: an energy function $J_\mathrm{g}: \R{\dimxi} \rightarrow \mathbb{R}_{\geq 0}$ assigns higher values to unsafe regions, and its gradient $\nabla_\vxi J_\mathrm{g}(\vxi)$ biases the sampled path away from obstacles. 
This yields a modified vector field $\tilde{v}_\zeta(\lambda, \vxi) = v_\zeta(\lambda, \vxi) - \gamma_\mathrm{g} \nabla_\vxi J_\mathrm{g}(\vxi)$, where $\gamma_g > 0$ controls the guidance strength. 
However, evaluating the flow-matching model is computationally expensive, and real-time systems can typically only afford a small number of integration steps~\cite{carvalho2025motion}. 
With a limited number integration steps, a large $\gamma_\mathrm{g}$ risks pushing the generative process out of distribution, while a small $\gamma_\mathrm{g}$ has negligible effect (see Fig.~\ref{fig:guidance}). 
Furthermore, updating the entire global plan for purely local environmental changes is computationally inefficient. 
To overcome these limitations, we leverage the MPPFC to locally re-plan the system's behavior, enforcing system dynamics and collision-avoidance constraints exclusively where needed (see Fig.~\ref{fig:scheme}).

At the beginning of task execution, we generate a batch of candidate paths from the generative model and select a collision-free path with respect to known obstacles. Rather than strictly requiring the flow-matching planner to output directly executable trajectories, our framework systematically delegates dynamic feasibility and real-time safety to the predictive controller. 
The selected path is incorporated into the MPPFC formulation in two complementary ways: it acts as a reference objective within the stage cost and defines the terminal region constraint (\ref{eq:OCP_termcon}). 
Specifically, the MPPFC constrains the system to return to and align with the reference path at the end of the prediction horizon. 
The exact realignment point is dynamically determined by the controller via the path dynamics and virtual input.

This hierarchical integration explicitly accounts for previously unknown static obstacles at runtime.\footnote{Moving obstacles further complicate the task, as additional models are needed to predict their behavior. Strategies for applying MPC in such scenarios have emerged, for example, in autonomous driving~\cite{Bruedigam2021,Bethge2023,Baltussen2025}.} 
Assuming the system is equipped with sensors (e.g., LiDAR), newly detected obstacles are convexified and incorporated into the safe set $\mathcal{X}_\mathrm{safe}$ as state constraints. 
Once the system safely bypasses an obstacle and it exits the sensing range, the associated constraints are removed to minimize the computational burden.

Due to the terminal alignment structure of the MPPFC, feasibility dictates that obstacle avoidance and path re-alignment must be achieved within a single prediction horizon. 
If an obstacle forces a significant deviation, strict terminal alignment may render the optimal control problem infeasible. To ensure robust collision-free operation, we conceptually decouple short-term collision avoidance from strict terminal alignment. 
While this can be achieved by softening the terminal constraint or employing a generative-guided terminal cost, we adopt a parallelized horizon-extension strategy. 
We solve multiple MPPFC instances simultaneously, each with an incrementally extended prediction horizon but a fixed number of collocation points. 
We then execute the solution from the instance with the shortest feasible horizon (see Fig.~\ref{fig:scheme}). 
Because a shorter horizon with fixed collocation points implies a smaller integration step size, it naturally yields higher control fidelity. 
The controller autonomously reverts to the shortest horizon once local feasibility is re-established.

For substantial deviations, event-triggered re-planning using the expert model is employed to improve adherence to the demonstrated global behavior.
Re-planning is triggered when the system exits a user-defined neighborhood around an evaded obstacle --- after the MPPFC has successfully ensured local safety. 
This allows the global planner to realign the overall trajectory with the expert distribution. 
Crucially, switching the reference path within the MPPFC operates seamlessly without requiring modifications to the underlying algorithmic procedure.

Ultimately, this systematic integration of generative planning and predictive control prioritizes real-time adaptability and robust empirical performance over conservative formal constraints. 
By combining the global geometric awareness of flow-matching with the strict local safety filtering of the MPPFC, the proposed strategy consistently produces collision-free behavior in practice, assuming the existence of a topologically viable alternative path, a standard and practical assumption for agile robotic platforms operating in task-oriented environments.

\subsection{Biasing the Planner Toward the MPPFC Solution}
\label{sec:planner_biasing}

A key component of the proposed method is re-planning the reference path during evasive maneuvers to improve adherence to demonstrated behaviors.
We achieve this by using the predictive controller’s collision-free solution, which incorporates updated local sensor and constraint information, to bias the generative planner toward locally safe trajectories, ensuring candidate paths remain consistent with the current system state.

Introducing this sampling bias in classical generative models, such as VAEs~\cite{kingma2013vae}, is not straightforward.
In denoising-diffusion-like methods, such as flow matching, this is easily integrated by applying noise steps to the previous MPC solution, followed by denoising steps.
This approach ensures the generated trajectories remain closer to the MPC prediction than sampling from standard Gaussian noise followed by multiple denoising steps.
A similar approach has been used in diffusion models for autonomous racing~\cite{ma2025constraint}.



Specifically, instead of starting the generative process from pure noise $\vxi_0 \sim \mathcal{N}(0, \mathbb{I})$, we propose initializing it from an intermediate step $\lambda = \bar{\lambda}$, with $0 < \bar{\lambda} < 1$. 
This requires a partially noisy sample $\vxi_{\bar{\lambda}} = (1-\bar{\lambda})\vxi_0 + \bar{\lambda} \vxi_1$, which depends on a clean sample $\vxi_1$.
Since the true $\vxi_1$ is generally unavailable, we use the MPC's latest prediction to obtain a local approximation $\hat{\vxi}_1$. 
This estimate is obtained by fitting a B-spline using (\ref{eq:fit_para}), providing a fast-to-compute, biased initialization. 
The generative process is then started with $(\vxi_{\bar{\lambda}}, \bar{\lambda})$ based on $\hat{\vxi}_1$, producing candidate paths that are locally aligned with the controller's current plan and likely maintain its collision-avoidance guarantees locally.

In addition to aligning the generative process with the MPC plan, the biasing provides notable computational advantages. 
By initializing the flow-matching model from an intermediate state informed by the MPC prediction, sampling typically requires fewer refinement steps and can therefore be performed more efficiently. 
Furthermore, by increasing the likelihood of obtaining a collision-free alternative reference path, the bias often reduces the magnitude of corrective actions required by the MPC. 
This, in turn, generally decreases the computational effort associated with solving the MPC problem.

\section{Simulation Experiments}
\label{sec:Simulation}

In this section, we evaluate our framework in simulated tasks.
The experiments are constructed to answer the following questions:
\begin{enumerate}
    \item What is the system's behavior in a navigation task with unknown obstacles?
    \item How do the MPC horizon expansion and re-planning strategies influence the success rate?
\end{enumerate}

\begin{figure}[t!]
    \centering
    \includegraphics[scale=0.4, trim={0.25cm 0.1cm 0.2cm 0.1cm},clip]{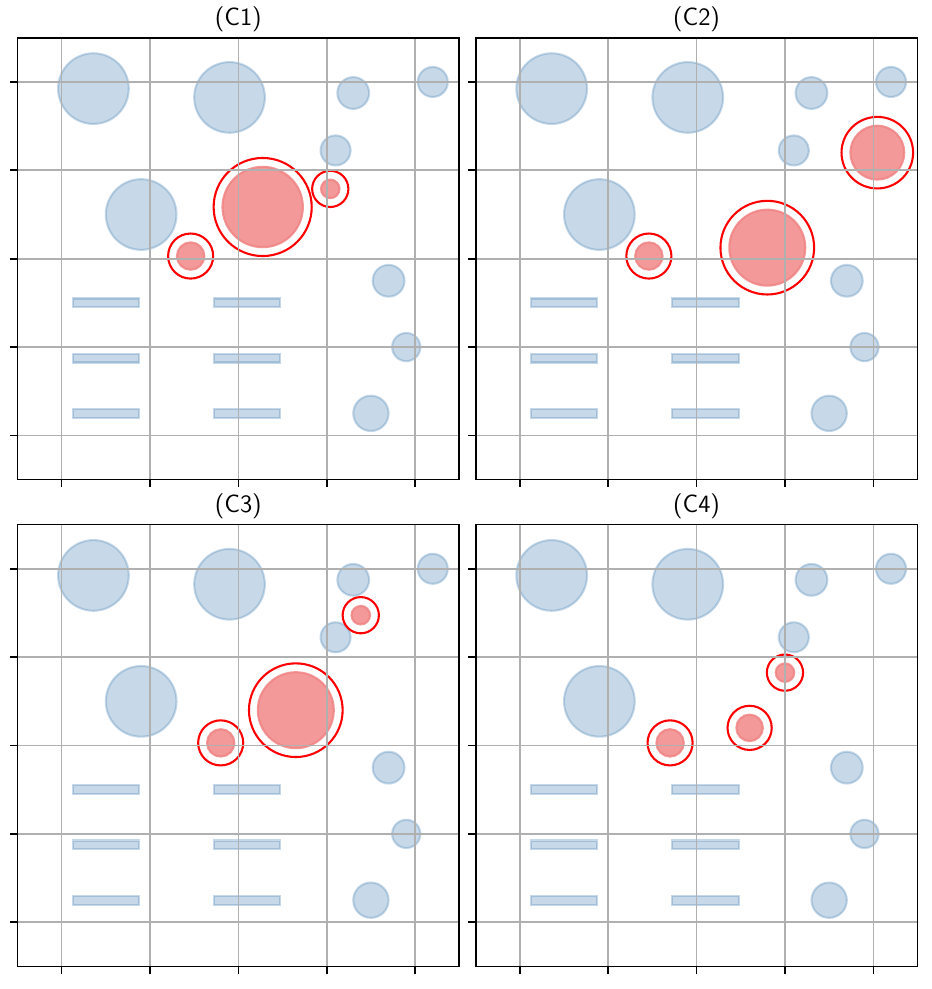}
    \vspace{-3mm}
    \caption{Simulation environments.
    Obstacles in blue are present during the demonstrations.
    During deployment, unknown objects in red might appear.    
    }\vspace{-3mm}
    \label{fig:env}
\end{figure}

\subsection{Environment, Data Collection, and Shortcut Model}

\begin{figure*}[t!]
    \centering
    \includegraphics[width=\textwidth]{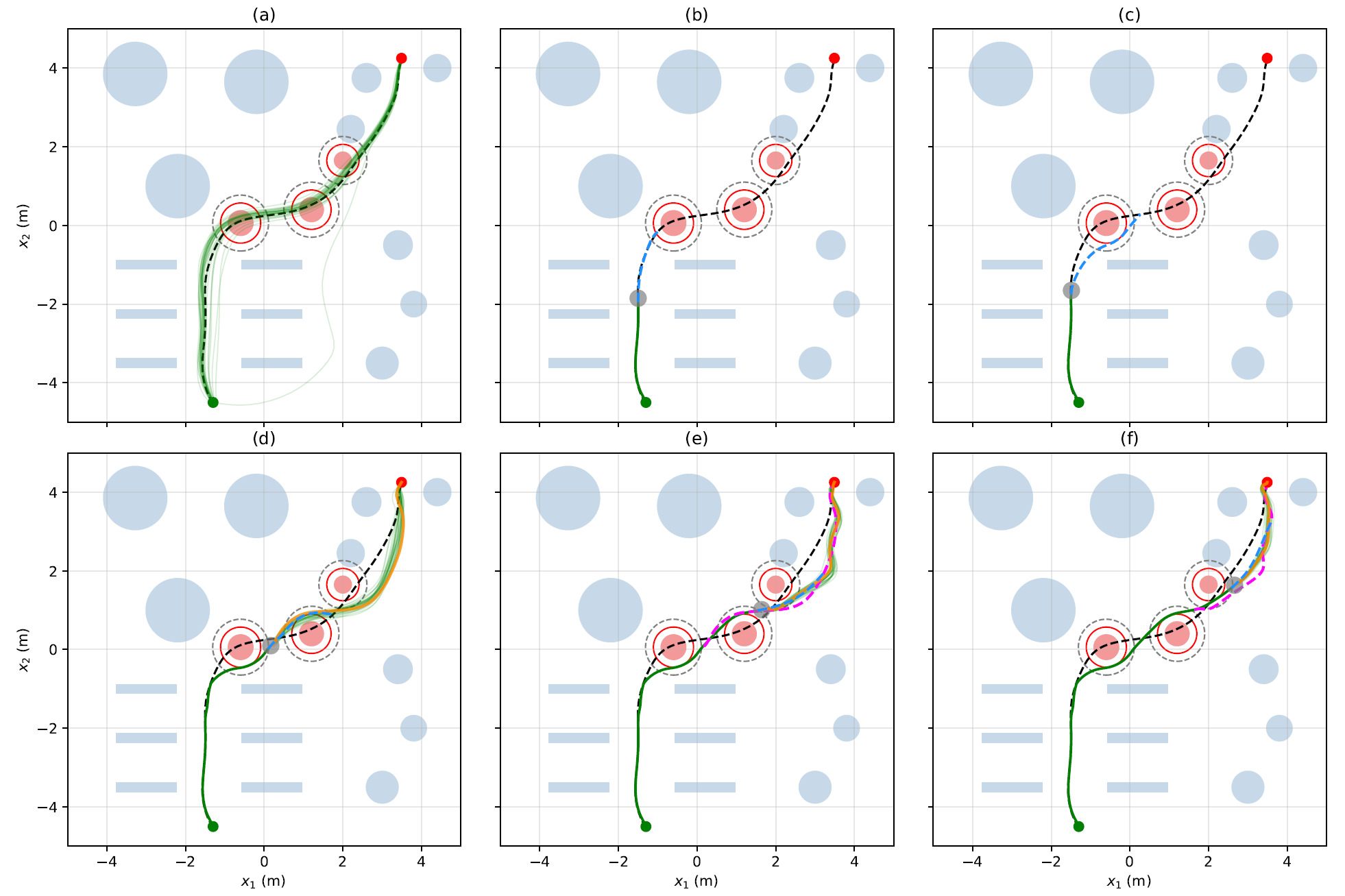}
    \vspace{-5mm}
    \caption{Simulation of a mobile robot navigating a cluttered environment. The robot (gray circle \graycircle) moves from a start (green circle \greencircle) to a goal position (red dot \redcircle). Known static obstacles are blue (\bluecircle), while previously unknown obstacles are red circles (\lightredcircle) with a red boundary indicating the robot's size margin (\redboundcircle). The generative planner produces candidate trajectories (solid light-green lines \solidlightgreen), from which a single path (dashed black line \dashedblack) is selected. The executed closed-loop path is dark green (\soliddarkgreen), and the current predicted trajectory is dashed light blue (\dashedlightblue). A yellow line (\solidyellow) denotes a newly chosen path from intermediate samples, while the dashed magenta line (\dashedmagenta) traces the intermediate path currently being followed (the prior yellow path). The dashed gray line (\dashedgray) defines the margin condition for sampling new paths.}\vspace{-3mm}
    \label{fig:simulation}
\end{figure*}

To train the shortcut model effectively, we require a diverse and informative dataset of collision-free paths. 
Such a dataset is constructed in a physically realistic MuJoCo simulation~\cite{todorov2012mujoco}, where a human operator navigates a mobile robot through a cluttered environment populated with various static obstacles (see Fig.~\ref{fig:simulation}). 
The operator guides the robot between multiple random start and goal positions, producing paths that, due to the non-deterministic nature of human behavior, exhibit a multimodal distribution.
The resulting dataset consists of approximately $1800$ demonstrations.
Our shortcut model is based on a diffusion transformer architecture as in~\cite{Peebles2023}.
The model is trained on an NVIDIA A100 GPU, and evaluated on an NVIDIA GeForce RTX 3090 GPU.

\subsection{Dynamics, Environments, and MPPFC Design}



We consider a mobile robot with dynamics
$\dot{x} = [u_1 \cos x_3,~ u_1 \sin x_3,~ u_2]^\top$,
where $x = [x_1, x_2, x_3]^\top$ denotes the state (planar position $(x_1, x_2)$ in $\SI{}{\m}$ and heading $x_3$ in $\SI{}{\rad}$), and the inputs are the translational velocity $u_1 \in [-1.5, 1.5]$ (in $\SI{}{\m~\s^{-1}}$) and the angular velocity $u_2 \in [-1.5, 1.5]$ (in $\SI{}{\rad~\s^{-1}}$). 
The output is $y = (x_1, x_2)$, and simulations run with a $50$ $\SI{}{\ms}$ sampling time. 



We consider four environments (C1–C4) (Fig.~\ref{fig:env}) containing identical static obstacles (blue) but varying configurations and sizes of previously unknown obstacles (red). 

The MPPFC instances are designed according to (\ref{eq:OCP}), with obstacles within the system’s sensing range defining the state constraint set $\mathcal{X}_\mathrm{safe}$ to enforce obstacle avoidance and ensure local reactivity.
We initially tested the MPC framework using a sampling-based approach, but found that the resulting inputs lacked smoothness, which negatively impacted path-following accuracy and traversal time.
Consequently, we focus on gradient-based implementations using CasADi \cite{Casadi}, which provide smoother control inputs and more reliable path-following performance.


We test three re-planning conditions: (P1) resampling every 10 closed-loop steps; (P2) resampling when a newly detected obstacle causes a collision in the MPC prediction\footnote{Obstacle detection is simulated via a LiDAR sensor.}; and (P3) resampling upon exiting a safety margin around an evaded obstacle.

\subsection{Closed-Loop Performance}

We first consider the environment C4 and the setting shown in Fig.~\ref{fig:simulation}.
We employ $K=15$ parallel controller instances with a base horizon length $T = \SI{1.0}{\s}$, an increment $\delta_t = 0.2$, and 20 equidistant collocation points along each horizon.
This results in a maximum horizon length of $\SI{3.8}{\s}$.
We employ the re-planning strategy P3, as proposed earlier.

We sample a batch of paths from the shortcut model, conditioned on a given start and desired goal position (Fig.~\ref{fig:simulation}a). 
All sampled paths are collision-free with respect to the obstacles encountered during demonstrations. 
From this batch, a path is randomly selected, and the parallel MPC instances compute the corresponding optimal inputs.
The solution from the instance with the shortest horizon that returns a feasible solution is then executed.



After some time, a previously unknown obstacle appears within the system's sensing range, causing the current reference path to be in collision (\ref{fig:simulation}b). 
To initiate an evasive maneuver, we include collision avoidance with the detected obstacle as a hard constraint in the MPC. 
However, adding this constraints can make the optimal control problem infeasible.
Specifically, the MPPFC must re-align with the path at the end of each prediction horizon, and depending on the horizon length, input limits, and other state constraints, this may no longer be achievable for any controller instance.


In such cases, the controller with the shortest horizon among the feasible instances is still selected and executed.
Moreover, whenever a new obstacle is added as a collision-avoidance constraint, we initialize the path parameter for the next horizon with $\theta(T) = 1$, corresponding to the end of the reference path.
This has been found beneficial, as it biases the optimizer toward solutions that prioritize significant path progress, increasing the likelihood that one controller instance remains feasible while planning to return to the reference path at the horizon's end (Fig.~\ref{fig:simulation}c).
If necessary, the prediction horizons of each controller instance can be iteratively extended until feasibility is recovered, depending on the available computation budget. 

When the robot exits a user-defined neighborhood around an evaded obstacle (illustrated as a circle in Fig.~\ref{fig:simulation}d, gray dashed), re-planning is triggered and a new batch of paths (light green) is sampled from the current position to the desired goal.
As described in Section~\ref{sec:planner_biasing}, we bias the planner by initializing it with a noisy version of the MPPFC output at $\bar{\lambda} = 0.75$, rather than from standard Gaussian noise.
As a result, the planner generates high-quality, locally collision-free candidate paths with only two integration steps.

From the newly generated paths, we select one that is collision-free with respect to both the obstacles in the current sensing range and those encountered during demonstration (yellow line in Fig.~\ref{fig:simulation}d) and deploy it in all controller instances.
If no such path exists, the framework continues without interruption, relying on the parallel MPC instances to handle local feasibility.
This procedure is repeated whenever a new, previously unknown obstacle appears (Fig.~\ref{fig:simulation}e,f), until the mobile robot reaches its destination.

\subsection{Validation using Ablation Studies}

\begin{figure}[t]
    \centering
    \includegraphics[scale=0.6, trim={0.2cm 0.15cm 0.1cm 0.2cm},clip]{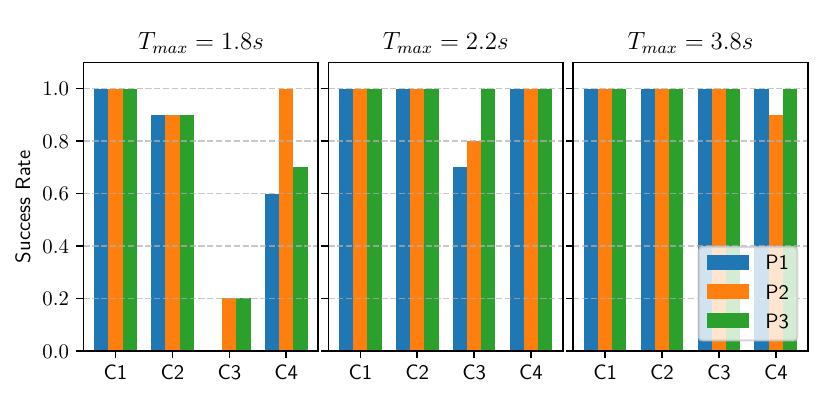}
    \vspace{-3.5mm}
    \caption{Success rates by obstacle configuration and re-planning strategy.}
    \label{fig:success_rate}
\end{figure}

\begin{figure}[t]
    \centering
    \includegraphics[scale=0.6, trim={0.1cm 0.0cm 0.1cm 0.1cm},clip]{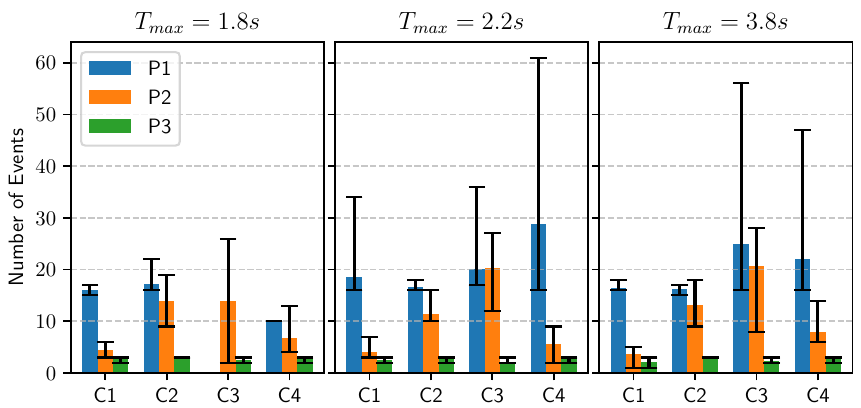}
    \vspace{-5.5mm}
    \caption{Number of re-planning events across obstacle configurations and re-planning strategies.}\vspace{-3mm}
    \label{fig:flow_evals}
\end{figure}

We conduct an ablation study to assess MPC horizon expansion and the re-planning strategies.
The obstacle configurations and re-planning conditions are evaluated across $K=\{5, 7, 15\}$ controller instances, corresponding to maximum prediction horizons $T_\mathrm{max}=\{ \SI{1.8}{\s}, \SI{2.2}{\s}, \SI{3.8}{\s} \}$, respectively.
Due to the stochasticity of the generative model, each scenario is simulated $10$ times.
Results for all tested combinations are shown in Fig.~\ref{fig:success_rate} in terms of success rates (successful trials over total trials). 
The number of re-planning events using the generative planner is reported in Fig.~\ref{fig:flow_evals}, considering only successful trials.

For $T_\mathrm{max} = \SI{1.8}{\s}$ ($K=5$), the controller is not always able to find a feasible solution across all scenarios, indicating that the prediction horizon is too short to provide consistently reliable performance. 
Increasing the horizon to $T_\mathrm{max} = \SI{2.2}{\s}$ ($K=7$) improves robustness, although obstacle configuration C3 remains challenging. 
With the longest horizon, $T_\mathrm{max} = \SI{3.8}{\s}$ ($K=15$), all tested combinations achieve a $100 \, \%$ success rate, except for P2 in C4, which achieves $90 \, \%$.
Notably, P3 achieves a $100 \, \%$ success rate across all configurations for both $T_\mathrm{max} = \SI{2.2}{\s}$ and $T_\mathrm{max} = \SI{3.8}{\s}$, while requiring fewer evaluations of the generative planner and thus improving computational efficiency without compromising performance.
Overall, the results indicate that the approach enables effective and efficient path planning and tracking in cluttered environments, maintaining collision-free behavior while outperforming naive baseline implementations.

\section{Conclusions}
\label{sec:Conclusions}

We proposed a framework that combines flow-matching-based generative planning, trained on expert demonstrations, with model predictive path-following control for adaptive navigation in cluttered environments. 
The generative planner provides diverse candidate paths, while the predictive controller incorporates onboard sensor information to adapt these paths online and ensure constraint satisfaction and collision-free execution, including in the presence of previously unseen obstacles. 
Event-triggered re-planning further couples path generation and control throughout task execution.
Future work will focus on incorporating richer contextual information, such as perception data, into path generation and extending the framework to dynamic environments through motion prediction.


\section*{ACKNOWLEDGMENTS}
During the preparation of this work, the authors used ChatGPT 5 and 5-Thinking by OpenAI to improve language and wording. The authors reviewed and edited the generated content and take full responsibility for the final publication.




\printbibliography


\end{document}